\documentclass{article}

   \usepackage[nonatbib,preprint]{neurips_2024}

\usepackage[utf8]{inputenc} 
\usepackage[T1]{fontenc}    
\usepackage{url}            
\usepackage{booktabs}       
\usepackage{amsfonts}       
\usepackage{nicefrac}       
\usepackage{microtype}      
\usepackage{xcolor}         
\usepackage{graphicx}
\usepackage{booktabs}
\usepackage{standalone}
\usepackage{subcaption}
\usepackage{amssymb}
\usepackage{pifont}
\usepackage{wrapfig}

\usepackage[accsupp]{axessibility}  

\usepackage{cuted}
\usepackage{capt-of}
\usepackage{hyperref}

\usepackage{orcidlink}

\definecolor{customgreen}{RGB}{0,128,0} 

\newcommand{\cmark}{\ding{51}}%
\newcommand{\xmark}{\ding{55}}%
\definecolor{darkblue}{RGB}{46,25, 110}

\title{EgoPet: Egomotion and Interaction Data from an Animal's Perspective}

\author{%
\normalsize\bf
  Amir Bar$^{1,2}$ \quad
  Arya Bakhtiar$^2$ \quad
  Danny Tran$^2$ \quad 
  Antonio Loquercio$^2$ \quad 
  Jathushan Rajasegaran$^2$ \\
  \quad\quad\quad\quad\normalsize\bf
  Yann LeCun$^3$ \quad 
  Amir Globerson$^1$ \quad 
  Trevor Darrell$^2$ \\[0.5em]
    \normalsize\quad\quad
    $^1$Tel Aviv University \quad 
    $^2$UC Berkeley \quad 
    $^3$New York University
}

\begin{document}

\maketitle

\begin{figure}[h]
\vspace{-5mm}
\centering
\includegraphics[width=1\textwidth]{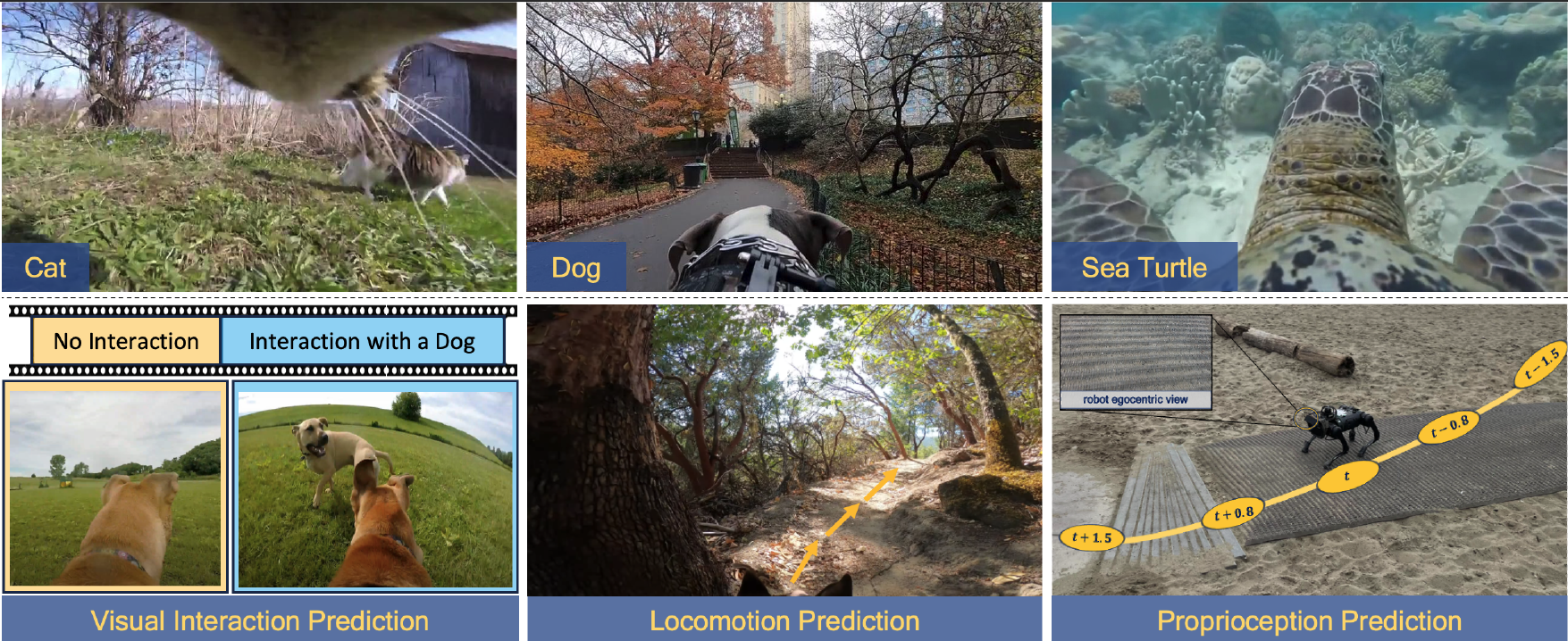}
\captionof{figure}{
  We present EgoPet, a novel animal egocentric video dataset to advance learning animal-like behavior models from video (top row). We propose three benchmark tasks on this dataset (bottom row). \emph{Visual Interaction Prediction} (VIP) and \emph{Locomotion Prediction} (LP) are designed to predict animals' perception and action behavior. Finally, \emph{Vision to Proprioception Prediction} (VPP) studies the utility of our dataset on the downstream task of robot locomotion in the wild. For all tasks, we find that models trained on EgoPet outperform those trained on previously available video datasets.}

\label{fig:teaser}
\end{figure}

\begin{abstract}
\vspace{-3mm}
Animals perceive the world to plan their actions and interact with other agents to accomplish complex tasks, demonstrating capabilities that are still unmatched by AI systems. To advance our understanding and reduce the gap between the capabilities of animals and AI systems, we introduce a dataset of pet egomotion imagery with diverse examples of simultaneous egomotion and multi-agent interaction. Current video datasets separately contain egomotion and interaction examples, but rarely both at the same time. In addition, EgoPet offers a radically distinct perspective from existing egocentric datasets of humans or vehicles.  We define two in-domain benchmark tasks that capture animal behavior, and a third benchmark to assess the utility of EgoPet as a pretraining resource to robotic quadruped locomotion, showing that models trained from EgoPet outperform those trained from prior datasets.~\footnote{Project page: ~\url{www.amirbar.net/egopet}}
\end{abstract}

\begin{figure*}[t]
    \centering
\resizebox{1\textwidth}{!}{\includegraphics{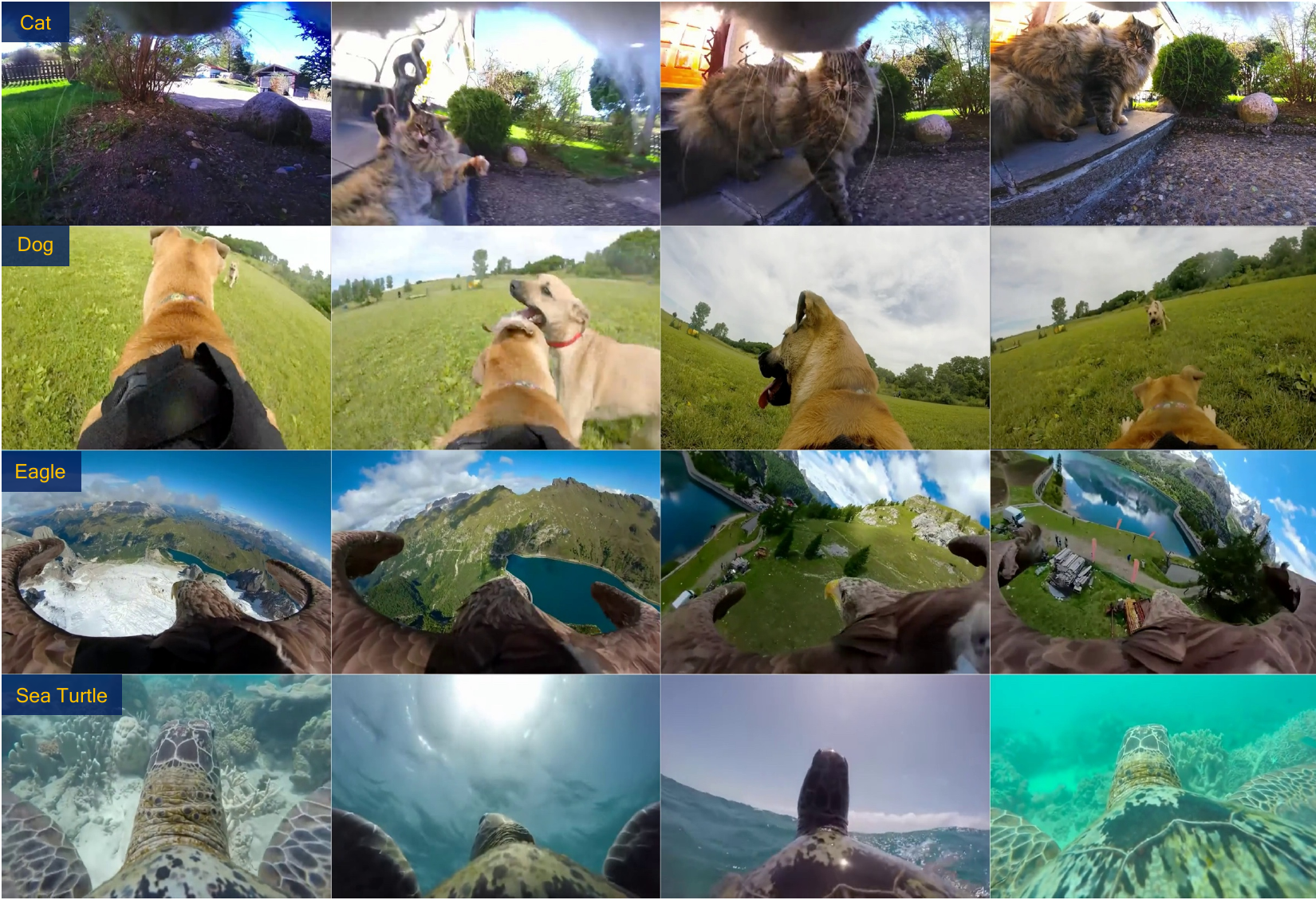}
}
 \caption{\textbf{EgoPet video examples}. Footage from the EgoPet dataset featuring four different animal experiences, each captured from an egocentric perspective at a distinct point in time.}   
 \label{fig:dataset-examples}
 \vspace{-3mm}
\end{figure*}

\section{Introduction}
\label{sec:intro}
Animals are intelligent agents that exhibit various cognitive and behavioral traits. They plan and act to accomplish complex goals and can interact with objects or other agents. Consider a cat attempting to catch a rat; this requires the cat to execute a precise sequence of actions with impeccable timing, all while responding to the rat's efforts to escape. 
 
Current Artificial Intelligence (AI) systems can synthesize high quality images~\cite{ramesh2021zero,rombach2022high}, generate coherent text~\cite{bubeck2023sparks,touvron2023llama}, and even code Python programs~\cite{chen2021evaluating}. But despite this remarkable progress, there are basic animal behaviors that are beyond the reach of current models. Recently, there has been a significant body of research in robotics aimed at learning policies for quadruped locomotion, and other basic actions~\cite{lee2020learning, kumar2021rma, agarwal2023legged, margolis2022rapid, shah2023vint, choi2023learning, miki2022learning, bajcsy2023learning}. However, we argue that a major limitation in advancing towards more complex systems is the availability of large-scale, real-world data.

To address this, we present EgoPet, a new web-scale dataset from the perspective of pets. EgoPet contains more than $84$ hours of video, including different animals like dogs, cats, eagles, turtles, and more. This video footage reveals the world from the eye of the pet as perceived in its day-to-day life, e.g., a dog going for a walk or entering a park, or a cat wandering freely around a farm. The video data was sourced from the internet and predominantly includes pet video, hence we have named the dataset EgoPet.

To measure progress in modeling and learning from animals, we propose three new tasks that aim to capture perception and action (see Fig.~\ref{fig:teaser}): Visual Interaction Prediction (VIP), Locomotion Prediction (LP), and Vision to Proprioception Prediction (VPP). Together with these tasks, we provide annotated training and validation data used for downstream evaluation.

The VIP task aims to detect and classify animal interactions and is inspired by human-object interaction tasks~\cite{shan2020understanding}. We temporally annotated a subset of the EgoPet videos with the start and end times of visual interactions and the object of the interaction category. The categories, which include person, cat, and dog, were chosen based on how commonly they occurred as objects (for all the categories refer to Supplementary Section \ref{sec:vip_details}).

The goal of the LP task is to predict the future $4$ second trajectory of the pet. This is useful for learning basic pet skills like avoiding obstacles or navigating. We extracted pseudo ground truth trajectories using Deep Patch Visual Odometry (DPVO)~\cite{teed2024deep}, the best-performing SLAM system for our dataset. We manually filtered inaccurate trajectories in the validation data to ensure high-quality evaluation.

Finally, in the VPP task, we study EgoPet's utility for a downstream robotic task: legged locomotion. Given a video observation from a forward-facing camera mounted on a quadruped robot, the goal is to predict the features of the terrain perceived by the robot's proprioception across its trajectory. Making accurate predictions requires perceiving the landscape and anticipating the robot controls. This differs from previous works on robot visual prediction~\cite{loquercio2023learning,sojka2023learning,margolis2023learning}, which require conditioning over current robot controls and are thus challenging to train at scale. To assess performance in this task, we gathered data utilizing a quadruped robodog. This data includes paired videos and proprioception features, which are then utilized for subsequent training and evaluation processes. 

We train various self-supervised models and evaluate how they perform downstream using a simple linear probing protocol. We make the surprising finding that pretraining on EgoPet yields better performance than pretraining on other, much larger video datasets like Ego4D~\cite{grauman2022ego4d} and Kinetics 400~\cite{kay2017kinetics}. This indicates the inadequacy of current datasets in studying animal-like physical skills.

Our contributions are as follows. First, we propose EgoPet, the first large-scale egocentric animal video dataset comprised of over $84$ hours of video footage to facilitate learning from animals. We propose three new tasks, including human-annotated data, and set an initial benchmark. The downstream results on the VPP task indicate that EgoPet is a useful pretraining resource for quadruped locomotion, and the benchmark results on VIP show that the proposed tasks are still far from being solved, providing an exciting new opportunity to build models that capture the world through the eyes of animals.
\section{Related Work}
\label{sec:rw}
Next, we delve into the related works surrounding video datasets, including notable research on both general video datasets of human and animals and those focusing specifically on egocentric video data.

\vspace{1.3mm}
\noindent\textbf{Video Datasets.} In recent years, a variety of video datasets have played an important role in video understanding tasks. In human action recognition, datasets like UCF101~\cite{soomro2012ucf101}, Charades-Ego~\cite{sigurdsson2018charades}, AVA~\cite{gu2018ava}, FineDiving~\cite{xu2022finediving}, and the Something-Something dataset~\cite{goyal2017something} provide comprehensive coverage of human activities, ranging from daily actions to specialized sports movements. Among these, Kinetics (K400)~\cite{kay2017kinetics} is particularly influential, advancing the study of human actions through a wide array of video clips. 

Other works aimed to collect data to study animals. These works include datasets such as the Animal Kingdom~\cite{ng2022animal}, which contains videos of various species, and MacaquePose~\cite{labuguen2021macaquepose}, which focuses on non-human primates. These datasets are instrumental for AI advancements in wildlife recognition and interpretation. AP-10K~\cite{yu2021ap} further augments this domain by providing a detailed collection of animal images for robust pose estimation. While sharing a similar motivation to our work, existing datasets on animal behavior rarely contain egocentric views and are therefore better suited to recognition problems than the animals' physical capabilities.  For autonomous driving and vehicle motion, datasets like the Berkeley DeepDrive~\cite{yu2020bdd100k,geiger2013vision} and KITTI~\cite{geiger2013vision} offer extensive insights into vehicle egomotion and environmental interactions. While these datasets enrich our understanding of motion, behavior, and interaction from a human-centric perspective, they offer limited insights into animal behavior.

\vspace{1.3mm}
\noindent\textbf{Egocentric Video Datasets.} Agents interact with the world from a first-person point of view, thus collecting such data has many applications from video understanding to augmented reality. In the past decade, many egocentric datasets were collected~\cite{fathi2012learning,damen2018scaling, sigurdsson2018charades,pirsiavash2012detecting,lee2012discovering}, with the majority of them focusing on human activities and object interactions in an indoor environment (e.g. kitchens). For example, Epic Kitchens~\cite{damen2018scaling, damen2020rescaling} is a large cooking dataset that takes place in $45$ kitchens across $4$ different cities, whereas Charades-Ego~\cite{sigurdsson2018charades} consists of $4{\small,}000$ paired videos of human actions in first and third person. Other datasets are more focused on conversation and social interactions~\cite{fathi2012social,ng2020you2me,northcutt2020egocom}. Existing datasets differ by the environments in which they are recorded, e.g. (outdoor vs. kitchens), whether they are scripted or not, and the number of videos. Recently, Ego4D~\cite{grauman2022ego4d}, a new comprehensive egocentric dataset was released. Different from previous datasets, it is more diverse (e.g., indoor and outdoor activities, different diverse geographical locations). However, while existing datasets focus on human and human skills, our focus is on animal agents which have more limited language and hand-object interactions. The most related egocentric dataset is DECADE~\cite{ehsani2018let} which consists of an hour of footage of a single dog, including joint locations annotations. Inspired by DECADE, EgoPet is a much larger web-scale dataset (84 hours) and much more diverse.

\section{The EgoPet Dataset}
\label{sec:egopet}
The EgoPet dataset is a unique collection of egocentric video footage primarily featuring dogs and cats, along with various other animals like eagles, wolves, turtles, sea turtles, sharks, snakes, cheetahs, pythons, geese, alligators, and dolphins (examples included in Fig.~\ref{fig:dataset-examples} and Suppl. Figure~\ref{fig:supp_dataset}). Together with the proposed downstream tasks and benchmark, EgoPet is a valuable resource for researchers and enthusiasts interested in studying animals from an egocentric perspective.

We begin with the motivation behind EgoPet and its connection to existing datasets in Section~\ref{sec:relation}. We then delve into the dataset's statistics in Section~\ref{sec:descriptive} and the collection process in Section~\ref{sec:datacollection}. 

\begin{figure*}[t]
    \centering
\includegraphics[width=\textwidth]{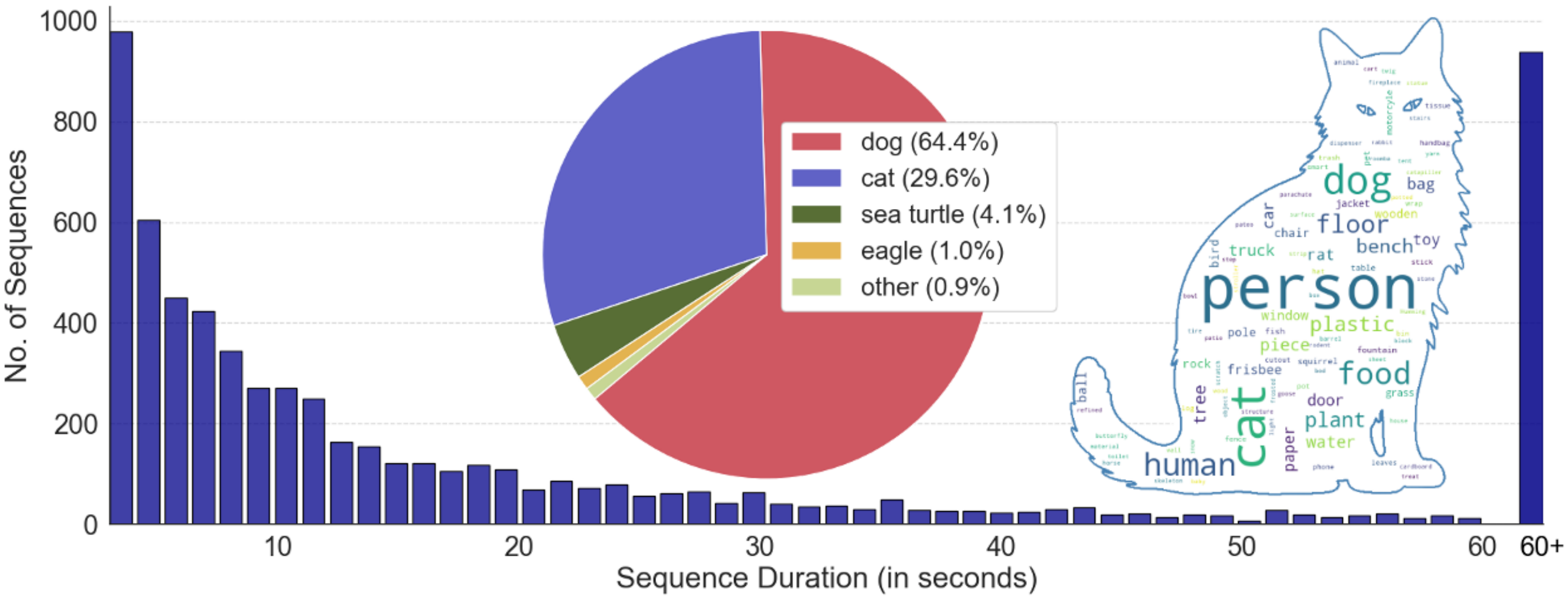}
    \caption{\textbf{Descriptive statistics.} The histogram depicting the length (in seconds) of EgoPet video sequences exhibits a long-tailed distribution, primarily skewed toward shorter segments of less than $30$ seconds. Collectively, videos featuring dogs and cats account for 94\% of the total duration, showcasing interactions with people, fellow cats and dogs, toys, and various objects.}
        \label{fig:descriptive}
\end{figure*}
\subsection{Relation to other datasets}
\label{sec:relation}
To provide a clearer understanding of EgoPet's significance, we compare it with various other datasets, considering factors such as total video duration, perspective (egocentric or non-egocentric), egomotion, the agents involved, and the presence of interaction annotations, which are crucial for intelligent agents. Refer to Table~\ref{tab:data-comparison} for more details. In terms of size, Ego4D~\cite{grauman2022ego4d} is the largest egocentric video dataset, and it centers on human activities, while the BDD100K~\cite{yu2020bdd100k} dataset includes both egocentric and egomotion elements but it focuses on autonomous driving. Differently, EgoPet focuses on animals, and pets in particular. Among animal video datasets, the DECADE~\cite{ehsani2018let} dataset provides an egocentric perspective from a dog's viewpoint, but it only records $1.5$ hours of video. EgoPet expands this vision by over $56$ times in volume and includes a variety of species and interactions.

\subsection{Descriptive Statistics}
\label{sec:descriptive}
The EgoPet dataset is an extensive collection composed of $6{\small,}646$ video segments distilled from $819$ unique videos. High level statistics are provided in Fig.~\ref{fig:descriptive}. These original videos were sourced predominantly from TikTok, accounting for $482$ videos, while the remaining $338$ were obtained from YouTube. The aggregate length of all video segments amounts to approximately $84$ hours, which reflects a substantial volume of data for in-depth analysis. In terms of video duration, the segments exhibit an average span of $45.55$ seconds, although the duration displays considerable variability, as indicated by the standard deviation of $192.19$ seconds. This variation underscores the range of contexts captured within the dataset, from brief encounters to prolonged interactions.

Breaking down the dataset by animal representation, cats and dogs constitute the majority, with $4{\small,}567$ and $1{\small,}905$ segments, respectively. This reflects the dataset's strong emphasis on common domestic animals while still covering less frequent but equally important species. Notably, the dataset includes segments featuring eagles ($66$), turtles ($31$), and a diverse group of other animals such as alligators, lizards, and dolphins, contributing to a rich collection of animal behaviors captured through an egocentric lens.

The camera positioning—where the recording device was attached to—also varies, the majority of segments were captured from cameras placed on the neck ($4{\small,}575$) and body ($1{\small,}817$). Fewer segments were recorded from cameras positioned on the head ($199$), shell ($36$), collar ($11$), and fin ($8$) offering a range of perspectives that can inform on how different mounting points might influence the perception of the environment from an animal's viewpoint.

\begin{table*}[t]
\centering

\resizebox{\linewidth}{!}{%
\begin{tabular}{l|ccccc}
\hline
Dataset	& Total Time (hours) & Egocentric & Egomotion & Agent & Interaction Annotations \\
\hline
BDD100K	 & $1{\small,}111$ & \cmark & \cmark & Cars &  \xmark  \\
Animal Kingdom &  $50$ & \xmark & \xmark & Animals & \xmark \\
EGO4D &	$3{\small,}670$ & \cmark & \xmark & Humans & \cmark \\
DECADE & $1.5$ & \cmark & \cmark & Dog & \xmark \\
\hline
EgoPet & $84$  & \cmark & \cmark & Animals & \cmark \\
\hline
\end{tabular}}
\caption{\textbf{Different video datasets}. We compare EgoPet to different datasets with respect to the total time (hours), whether the videos are in first person view (egocentric) with the focus on egomotion, the agent type, and whether agent interaction annotations are available. EgoPet is the first large scale animal dataset that is both ego centric and contains interaction annotations. It is also over $56$ times larger than the previous similar dataset DECADE~\cite{ehsani2018let}.}
\label{tab:data-comparison}
\end{table*}
\subsection{Data Acquisition}
\label{sec:datacollection}

\noindent\textbf{Collection Strategy.} To collect the dataset, we manually searched for videos using a large set of queries on YouTube and TikTok. For example, ``egocentric view'', ``dog with a GoPro'', and similar phrases related to first-person animal perspectives. This led to scraping a vast pool of footage showcasing animals, primarily dogs and cats, wearing wearable cameras, allowing for an egocentric point of view. In pursuit of a broader video selection, our efforts extended to individual channels and authors known for their thematic consistency in publishing egocentric animal footage. This approach allowed us to tap into niche communities and content creators, yielding a wide variety of egocentric videos beyond the reach of generic search terms.

\vspace{1.3mm}
\noindent\textbf{Dataset Refinement.}
A meticulous annotation process was carried out to ensure the dataset's quality. A human annotator reviewed the collected videos to confirm that they were from an egocentric point of view. Non-egocentric or irrelevant segments were carefully removed. 

All videos were adjusted to a frame rate of $30$ frames per second and resized to 480p on the shortest side while maintaining the original aspect ratio. The videos were then segmented into discrete clips, during which any non-egocentric footage was removed. The final dataset consists of segments of at least three seconds, ensuring sufficient context for each interaction.

\section{EgoPet Tasks}
\label{sec:tasks}
In order to allow quantitative comparisons of animal-prediction approaches, we next define several prediction tasks on the EgoPet dataset. We provide annotated datasets based on these tasks, which will allow effective benchmarking of different approaches. 

\subsection{Visual Interaction Prediction (VIP)}
\noindent\textbf{Motivation.} Human activities such as actions and interactions from an egocentric viewpoint have been previously explored in various datasets mostly focusing on activity recognition \cite{kazakos2019epic, li2021ego, zhou2015temporal}, human-object interactions \cite{damen2016you, cai2016understanding, nagarajan2019grounded, damen2018scaling}, and social interactions \cite{fathi2012social, ng2020you2me, yonetani2016recognizing}. Inspired by these works, we focus on animal interactions with other agents or objects, and for simplicity, we only consider visual interactions. Observing interactions through an egocentric perspective offers insights into how animals navigate their world, how they communicate with other beings, and how their physical movements correlate with environmental stimuli. Being able to identify interactions is a core task in computer vision and robotics with practical applications in designing systems that can operate in dynamic, real-world settings. 

\vspace{1.3mm}
\setlength\intextsep{0pt}
\begin{wrapfigure}{tr}{0.5\textwidth}
    \centering
\includegraphics[width=1\linewidth]{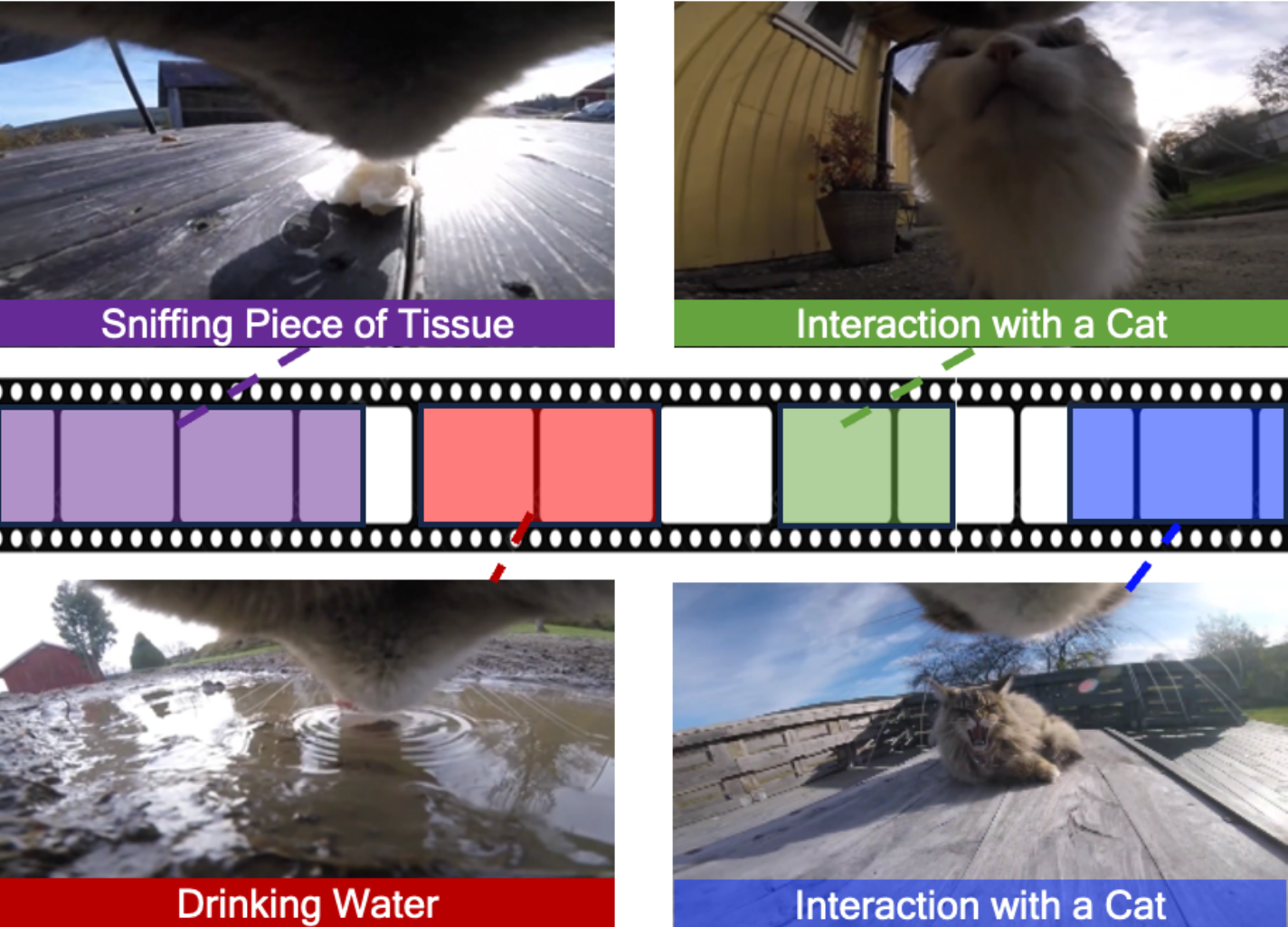}
    \caption{\textbf{Vision to Interaction Prediction task.} The figure illustrates the process of annotating a single video, identifying and categorizing different interactions experienced by a cat, with each segment of the timeline reflecting a unique type of interaction within the animal's environment.}
        \label{fig:VIP_2}
\label{fig:vip_fig}
\vspace{-2mm}
\end{wrapfigure}
\noindent\textbf{Task Description.} The input for this task is a video clip from the egocentric perspective of an animal. The labels are twofold: a binary label indicating whether an interaction is taking place or not, and a categorical label describing the object of the interaction. This binary label simplifies the vast range of potential interactions into a manageable form for the model, while the identification of the interaction object adds a layer of specificity necessary for understanding the context of the interaction.

In the context of the EgoPet dataset, a ``visual interaction'' is defined as a discernible event where the agent—typically an animal such as a dog or cat—demonstrates clear attention to an object or another agent within its environment. This attention may be manifested through physical contact, proximity, orientation, or vocalization (such as barking or making sounds) toward the object of the interaction which can be an object or agent. The fundamental criterion for a visual interaction is the presence of visual evidence within the video that the agent is engaged with, or reacting to a particular stimulus. Aimless movements, such as wandering without a clear target or displaying alertness without a specific focus, are not labeled as visual interactions.

\vspace{1.3mm}
\noindent\textbf{Annotations.} The data labeling process for marking interactions involved a meticulous analysis of the video content, which resulted in the annotation of $1449$ subsegments (see Fig.~\ref{fig:vip_fig}). Two human annotators were trained to identify and timestamp the start and end of an interaction event. The outcome of this process is a richly annotated dataset of $805$ subsegments where no interaction occurs (''negative subsegments'') and $644$ positive interaction subsegments that capture a wide range of $17$ distinct interaction objects such as person, cat, and dog. The subsegments were then split into train and test sets. This leaves us with $754$ training subsegments and $695$ test subsegments for a total of $1{\small,}449$ annotated subsegments. To see the full annotation process refer to Suppl. Section~\ref{sec:vip_details}. 

\subsection{Locomotion Prediction (LP)}
\noindent\textbf{Motivation.}
Planning where to move involves a complex interplay 
\setlength\intextsep{0pt}
\begin{wrapfigure}{t}{0.4\textwidth}
\vspace{-6mm}
    \centering
    \includegraphics[width=1\linewidth]{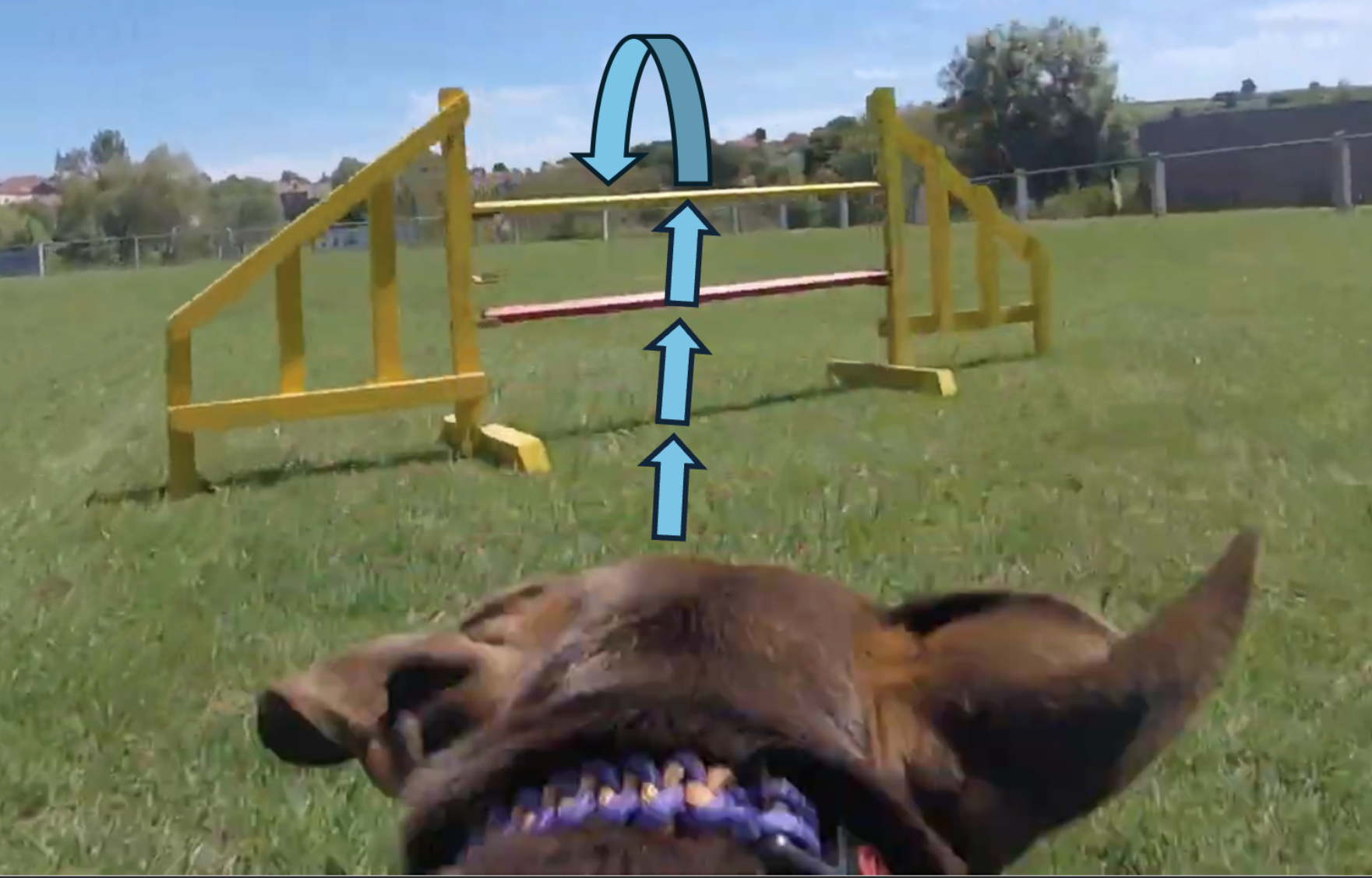} 
    \caption{\textbf{Locomotion Prediction task.} A dog navigates an agility course, highlighting the concept of locomotion prediction by anticipating its forward and upward trajectory to clear the obstacle.}
    \label{fig:lp_2}
\vspace{-4mm}
\end{wrapfigure}
of both perception and foresight. It requires the ability to anticipate potential obstacles, consider various courses of action, and select the most efficient and effective strategy to achieve a desired goal. EgoPet contains examples where animals plan a future trajectory to achieve a certain goal (e.g., a dog following its owner. See Fig.~\ref{fig:lp_2}).

\vspace{1.3mm}
\noindent\textbf{Task Description.} 
Given a sequence of past $m$ video frames $\{x_i\}^{t}_{i=t-m}$, the goal is to predict the unit normalized future trajectory of the agent $\{v_j\}^{t+k}_{j=t+1}$, where $v_j\in \mathbb{R}^{3}$ represents the relative location of the agent at timestep $j$. We predict the unit normalized relative location due to the scale ambiguity of the extracted trajectories. In practice, we condition models on $m=16$ frames and predict $k=40$ future locations which correspond to $4$ seconds into the future.

\vspace{1.3mm}
\noindent\textbf{Annotations.} To obtain pseudo ground truth agent trajectories, we used Deep Patch Visual Odometry (DPVO~\cite{teed2024deep}), a system for monocular visual odometry that utilizes sparse patch-based matching across frames.  This system largely outperformed other open-source SLAM systems in terms of convergence rate and qualitative accuracy in our experiments.

Given an input sequence of frames, DPVO returns the location and orientation of the camera for each frame. To obtain training trajectories, we feed videos with a stride of $5$ to DPVO. To ensure high-quality evaluation, we feed validation videos with strides of $5$, $10$, and $15$ into DPVO and evaluate the quality of the trajectories manually. Specifically, two human annotators were trained to evaluate the trajectories from an eagle's eye view (XZ view) and determine the best matching trajectory, if any, to the video. This left us with $6{\small,}126$ annotated training segments and $249$ validation segments.



\subsection{Vision to Proprioception Prediction (VPP)}
\begin{wrapfigure}{t}{0.5\textwidth}
    \centering
    \includegraphics[width=1\linewidth]{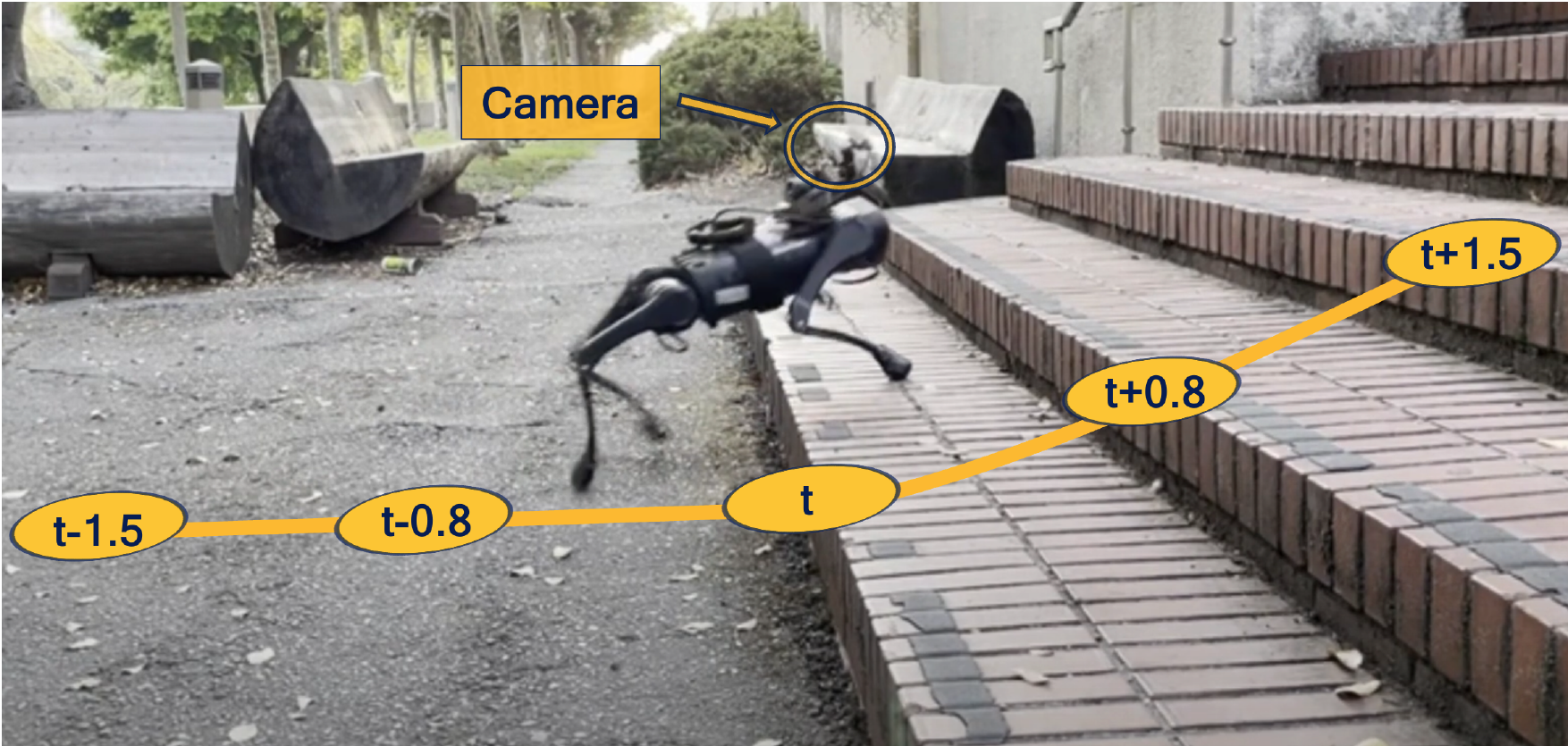}
    \caption{\textbf{Vision to Proprioception Prediction task.} This figure showcases the quadruped robot as it is about to transition from flat ground to climbing steep stairs, illustrating one of the unique terrain environments encountered during the collection of visual and proprioceptive data at annotated time intervals for VPP training.}
    \label{fig:vpp_2}
\end{wrapfigure}

\noindent\textbf{Motivation.} Understanding animal behavior could be instrumental to several robotics applications. To demonstrate the value of our dataset for robotics, we propose a task based on the problem of vision-based locomotion. Specifically, the task consists of predicting the parameters of the terrain a quadrupedal robot is walking on (see Fig.~\ref{fig:vpp_2}). As shown in multiple previous works on locomotion~\cite{loquercio2023learning,margolis2023learning,karnan2023self,bednarek2019touching, bednarek2019robotic, sojka2023learning}, accurate prediction of these parameters is correlated with improved performance in locomotion. Intuitively, the EgoPet data closely resembles the video captured by a quadruped robot since a camera mounted on a pet is approximately at the same location as the camera mounted on the robot. In addition, the task of walking is highly represented in the dataset.

\vspace{1.3mm}
\setlength\intextsep{0pt}
\noindent\textbf{Task Description.} 
The parameters we would like to predict are the local terrain geometry, the terrain's friction, and the parameters related to the robot's walking behavior on the terrain, including the robot's speed, motor efficiency, and high-level command. The exact identification of these parameters is generally impossible~\cite{kumar2021rma}: two terrains could have different combinations of parameters but ``feel" the same to an agent. For example, walking on sand and mud could similarly affect the robot's proprioception, even though their properties differ. Therefore, similarly to previous work~\cite{kumar2021rma,lee2020learning, loquercio2023learning}, we aim to predict a latent representation $z_t$ of the terrain parameters.
This latent representation consists of the hidden layer of a neural network trained in simulation to encode ground-truth terrain parameters. This neural network is trained end-to-end with an action policy on locomotion using reinforcement learning.
The task consists of predicting the latent terrain's representation at different time intervals from a sequence of frames. Our setup closely follows the one in~\cite{loquercio2023learning}. Specifically, we add to EgoPet data collected with a quadrupedal robot in multiple outdoor environments with different terrain characteristics, e.g., sand or grass.
%

\vspace{1.3mm}
\noindent\textbf{Dataset and Annotations.}
To collect the dataset, we deployed the walking policy of~\cite{loquercio2023learning} on a Unitree A1 robot dog in three environments: an office, a park, and a beach. We collect approximately 20 minutes of walking data in these environments, which are used exclusively for evaluation.  For training, we use the data from~\cite{loquercio2023learning}, which contains 120 thousand frames, corresponding to a total walking time of approximately 2.3 hours. Each environment has different terrain geometries, including flats, steps, and slopes. Each sample contains an image collected from a forward-looking camera mounted on the robot and the (latent) parameters of the terrain below the center of mass of the robot $z_t$ estimated with a history of proprioception. See~\cite{loquercio2023learning} for details about the annotation procedure.

The final task consists of predicting $z_t$ from a history of images. We generate several sub-tasks by predicting the future terrain parameters $z_{t+0.8}, z_{t+1.5}$ and the past ones $z_{t-0.8}, z_{t-1.5}$. These time intervals were selected to differentiate between forecasting and estimation. The further the prediction is in the future or the past, the harder the task is. The input images might contain little information, or not at all, about the terrain at these times. Therefore, inferences based on the context are required. For example, one can predict the presence of a step in front of the robot from the shadow it casts on the terrain.

We divide the newly collected data into three test datasets: the first is in-distribution, featuring terrains and lighting conditions similar to the training data. The second dataset is out of distribution since it is captured with different lighting conditions, i.e., at night, but in environments with the same features as the training data. Finally, the third dataset contains sandy environments, which the robot has not encountered during training.
\section{Evaluation Benchmark}

Our goal in the experiments is to establish initial performance baselines on the EgoPet tasks. For the VPP task, we hypothesize that EgoPet is a more useful pretraining resource compared to other datasets. We evaluate different pretrained models and compare their performance on the VIP, LP, and VPP tasks. We adopt a simple linear probing protocol, where we freeze the model weights and for each task train only a linear layer to predict the output. For evaluation, we use the following models publicly released checkpoints, typically trained on IN-1k or K400, unless stated otherwise.

\vspace{1.3mm}
\noindent\textbf{MAE} \cite{he2022masked} are trained by masking random patches in the input image and reconstructing the missing pixels through an asymmetric encoder-decoder architecture. By masking a high proportion (e.g., $75$\%) of the input image. In our experiments we use an MAE model pretrained on IN-1k.

\vspace{1.3mm}
\noindent\textbf{MVP} \cite{xiao2022masked} uses a similar model as in MAE, but trains MAE on a mixture of egocentric datasets which we refer to as Ego Mix, a combination of Epic-Kitchens~\cite{damen2018scaling}, 100DOH~\cite{shan2020understanding}, Ego4D~\cite{grauman2022ego4d}, and Something-Something~\cite{goyal2017something}.

\vspace{1.3mm}
\noindent\textbf{DINO} \cite{caron2021emerging} is trained with a student-teacher architecture over pairs of augmented images by encouraging invariance to the image augmentations. The teacher's output is centered, and both networks' normalized features are compared using a cross-entropy loss. The stop-gradient operator ensures gradient propagation only through the student, and teacher parameters are updated using an exponential moving average (ema) of the student parameters. 

\vspace{1.3mm}
\noindent\textbf{iBOT} \cite{zhou2021ibot} is trained similarly to DINO, but adds an auxiliary MIM loss by predicting image patch representation of a learned online tokenizer. 

\vspace{1.3mm}
\noindent\textbf{VideoMAE} \cite{tong2022videomae} is an extension of MAE for video pre-training. Different from MAE, it utilizes an extremely high masking ratio ($90$\% to $95$\%) and tube masking as opposed to random masking.

\vspace{1.3mm}
\noindent\textbf{MVD} \cite{wang2023masked} is a masked feature modeling framework for self-supervised video representation learning. Learning the video representations involves distilling student model features from both video and image teachers. We train MVD variants on Ego4D and EgoPet, using VideoMAE (K400) and MAE (IN-1k) as video and image teachers.

\vspace{1.3mm}
\noindent\textbf{Implementation Details.} For all models, we used the ViT-B model with patch size 16 since it was available across all methods. For the VIP task, we train all image and video models for $10$ epochs. Video models represent video clips of $2$ seconds using $8$ input frames ($4$ Hz) and image models use one (the middle) frame. For the VPP task, we train all models for $50$ epochs. Video models were trained with varying amounts of frames in $4$ Hz. For the LP task, we train all models for $15$ epochs. Video models were trained with $16$ input frames ($30$ Hz) and image models used one frame (the last). In our LP experiments, we only use cat and dog segments, segments long enough, and $25\%$ of the training data. This left us with $1${\small,}$129$ training segments and $167$ validation segments. During the linear probing training phase, we do not apply any image augmentations. All the other hyperparams follow MAE and MVD linear probing recipes for image and video models respectively.

\begin{table*}[t]
\centering
\begin{tabular}{llcccccccc}
\hline
Model & Dataset &\multicolumn{2}{c}{Interaction Prediction} & \multicolumn{2}{c}{Object Prediction} \\
& & Accuracy & AUROC & Top-1 Acc & Top-3 Acc \\
\hline
MAE & IN-1k & 62.34 & 69.41 & 35.02 & 61.37 \\
MVP & Ego Mix & 65.47 & 68.12 & 33.57 & 59.21 \\
DINO & IN-1k & 65.16 & 73.38 & 37.18 & 60.65 \\
iBOT & IN-1k & 65.16 & 73.50 & \textbf{37.55} & 58.12 \\
\hline
VideoMAE & K400 & 61.56 & 66.22 & 29.24 & 54.87 \\
MVD & K400 & 65.63 & 70.35 & 35.38 & 62.45 \\
MVD & Ego4D & 64.84 & 70.15 & 33.57 & 62.45 \\
MVD & EgoPet & \textbf{68.44} & \textbf{74.31} & 35.74 & \textbf{64.62} \\
\hline
\end{tabular}
\caption{\textbf{Visual Interaction Prediction linear probing results.} We report models Interaction Prediction Accuracy and AUROC, as well as Object Prediction Top-1 and Top-3 Accuracy.}
\label{tab:interaction}

\end{table*}

\section{Results}
In this section, we report initial baseline results from applying a range of models to the VIP, LP, and VPP tasks. Taken together, these results underline the interesting observation that current large video datasets used for pretraining are not diverse enough to perform well across all the EgoPet downstream tasks. For example, pretraining on K400 is better than Ego4D for VIP but worse on VPP. Furthermore, by pretraining on EgoPet, we observe improved downstream performance on the VPP task compared to other models.

\subsection{Visual Interaction Prediction}
\begin{wrapfigure}{r}{0.35\textwidth}
\vspace{-6mm}
  \centering
    \includegraphics[width=\linewidth]{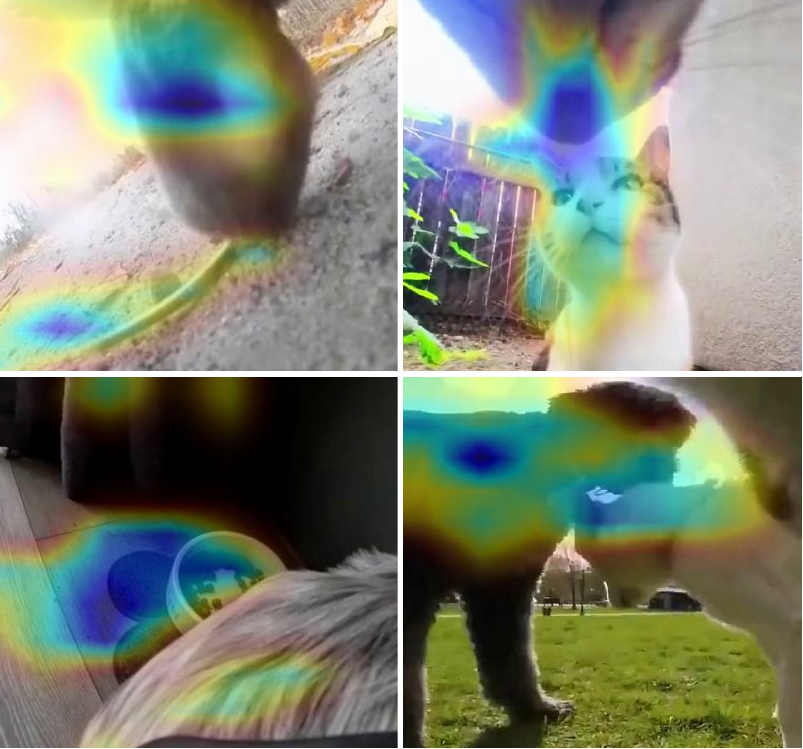}
    \caption{VIP Grad-CAM~\cite{selvaraju2017grad} visualization.}
    \label{fig:gradcam}
\end{wrapfigure}
The results in Table~\ref{tab:interaction} show that models trained on EgoPet achieve improved performance compared to K400 or Ego4D both on interaction prediction and object prediction. Compared to image based models like iBOT, MVD trained on EgoPet performs better on Top-3 Acc but worse on Top-1. This is likely due to the diversity of objects appearing in IN-1k, an image recognition dataset. Compared to other video models, MVD (EgoPet) performs better. To obtain more insight into what models focus on in this task, we apply Grad-CAM~\cite{selvaraju2017grad} on our MVD EgoPet interaction classifier. Fig.~\ref{fig:gradcam} shows the corresponding heatmaps and can be seen to focus on the rat (top-left), and another dog (bottom-right). In these cases, the model seems to be attending to the object of interaction. 

\begin{figure}
\centering
\includegraphics[width=0.9\textwidth]{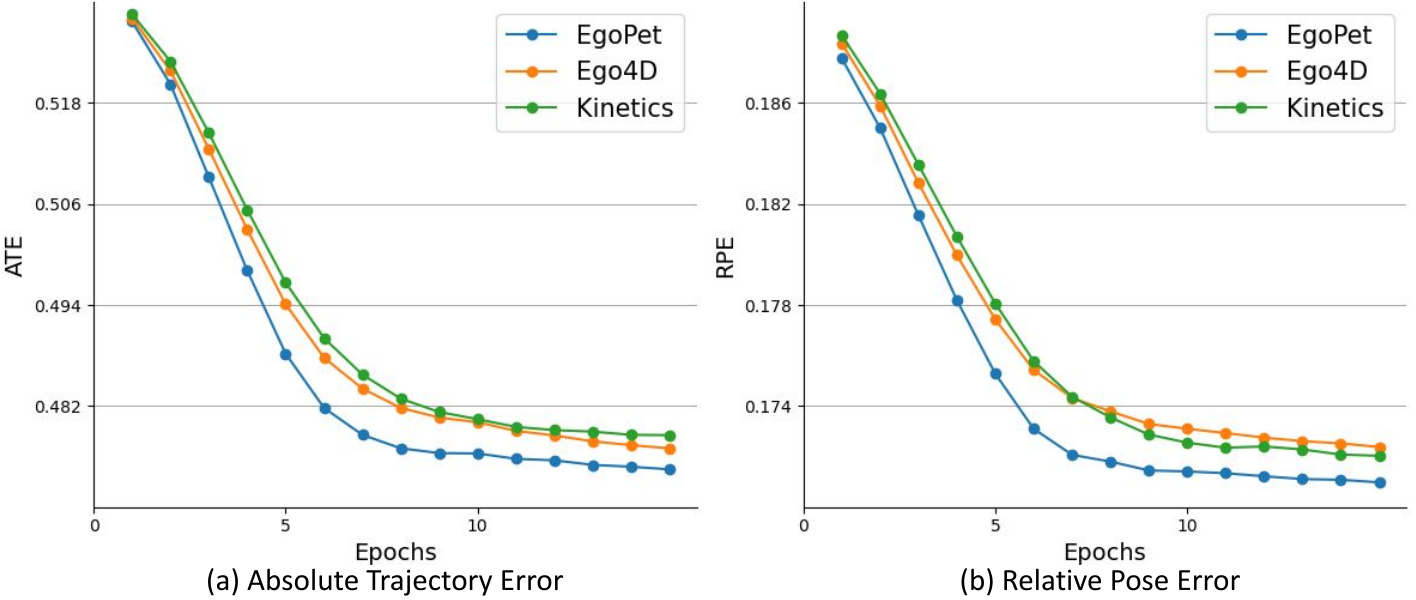}
\caption{
\textbf{Locomotion Prediction (LP) linear probing results.} We report the validation ATE and RPE as a function of the epoch during training, comparing the impact of various datasets (Kinetics, Ego4D, EgoPet). Models trained on EgoPet perform better than models trained on other datasets. See Supplementary Table~\ref{tab:planning} for the full results. }
\label{fig:lp_results}
\end{figure}

\subsection{Locomotion Prediction}
For this task, we evaluated models based on their predicted unit motions 40 timesteps into the future, corresponding to 4 seconds. We form trajectories from these predicted motions and compute the RMSE of the Absolute Trajectory Error (ATE) and Relative Pose Error (RPE) metrics against the ground truth trajectories. ATE and RPE are commonly employed metrics for evaluating systems such as SLAM and visual odometry \cite{zhao2022particlesfm, teed2024deep, teed2021droid, campos2021orb}. ATE first aligns the ground truth with the predicted trajectory, and then computes the absolute pose difference. RPE measures the difference between the predicted and ground truth locomotion \cite{sturm2012benchmark}.

The results in Fig.~\ref{fig:lp_results} indicate that models trained on EgoPet perform better than Ego4D and K400 and that the Ego4D model performed second best, possibly due to also being egocentric data. The full results in Supplementary Table~\ref{tab:planning} indicate that video models perform much better than image models as a whole, which we speculate is due to better modeling of the agent’s velocity and acceleration as well as the motion of other agents.

\subsection{Vision to Proprioception Prediction}
Table \ref{tab:vpp_res} provides the result of the VPP task. It can be seen that using EgoPet data leads to lower errors on this task. Additionally, the results show that using additional past frames as context helps and that video models outperform image models. Within image models, MVP performs better than MAE, likely because it was trained on egocentric data, and iBOT performs best. Within video models, MVD trained on EgoPet achieves lower mean squared error loss compared to the same model trained with K400 or Ego4D, and lower error compared to all image models. We speculate that MVD (EgoPet) outperforms all models because compared to other datasets the EgoPet videos are more similar to videos captured by a forward-facing camera mounted on a quadruped robodog. We provide the full results in the Supplementary Table~\ref{tab:vpp_supp}.

\begin{table}[t]
\centering
\begin{tabular}{llcccc}
\hline
Model & Dataset & Past $(t-k)$& Present $(t)$ & Future $(t+k)$ \\
\hline
\small{\textbf{1 Frame}} \\
MAE & IN-1k & 0.360 &0.280 &0.314 \\
MVP & Ego Mix &  0.357 &0.273 & 0.308\\
DINO & IN-1k &  0.354 &0.275 &0.304 \\
iBOT & IN-1k & 0.350 &0.278 &0.304 \\
\hline

\hline
\small{\textbf{4 Frames}} \\

MVD & K400 & 0.286 &0.197 &0.262 \\
MVD & Ego4D &  0.261 &0.224 &0.261  \\
MVD & EgoPet &  0.256 &0.203 & \textbf{0.246} \\
\hline
\small{\textbf{8 Frames}} \\
MVD & K400 & 0.217 &0.196 &0.252 \\
MVD & Ego4D & 0.208 &0.192 &0.249 \\
MVD & EgoPet & \textbf{0.204} & \textbf{0.184} &0.253 \\
\hline
\end{tabular}
\vspace{3mm}
\caption{\textbf{Vision to Proprioception Prediction (VPP) linear probing results}. We report the mean squared error loss. Models trained on EgoPet perform better than models trained on other datasets. See Supplementary Table~\ref{tab:vpp_supp} for the full results.}
\label{tab:vpp_res}
\end{table}

\section{Limitations} EgoPet is a video dataset, and as such it primarily contains visual and auditory signals. However, animals interact with their environment using a multitude of senses, including smell and touch. The absence of these sensory modalities in our dataset and model may lead to a partial or skewed understanding of animal behavior and intelligence. Animal behavior is highly complex and influenced by a myriad of factors, including instinct, learning, environmental stimuli, and social interactions. Our tasks, while effective in capturing certain aspects of behavior, may not fully encapsulate the depth and complexity of animal interactions and decision-making processes. Further research is needed to develop more sophisticated tasks and models that can account for complex behavioral patterns. 

\section{Conclusion}

We present EgoPet, a new comprehensive animal egocentric video dataset. Together with the proposed downstream tasks and benchmark, we believe EgoPet offers a testbed for studying and modeling animal behavior. Our benchmark results demonstrate that interaction prediction is far from being solved which provides an exciting opportunity for future research and modeling animal egocentric agents. Furthermore, the results demonstrate that EgoPet is a useful pretraining resource for downstream robotics locomotion tasks. Future works can include broadening the tasks to integrate more sensory inputs like audio, thereby creating a richer and more holistic understanding of the animal behavior.

{\bf Acknowledgements:}
We thank Justin Kerr for helpful discussions. Many of the figures use images taken from web videos. For each figure, we include the URL to its source videos in the Suppl. Section~\ref{suppl:credits}. This project has received funding from the European Research Council (ERC) under the European Unions Horizon 2020 research and innovation programme (grant ERC HOLI 819080). Prof. Darrell’s group was supported in part by DoD including DARPA's LwLL and/or SemaFor programs, as well as BAIR's industrial alliance programs.


%
%
\bibliographystyle{splncs04}
\bibliography{main}

\clearpage
\setcounter{page}{1}
\section*{Supplementary Material}
We provide additional information about the EgoPet dataset, annotation process and full quantitative results. 

\section*{Dataset}
\begin{figure}[h]
\centering
\vspace{3mm}
\includegraphics[width=1\linewidth]{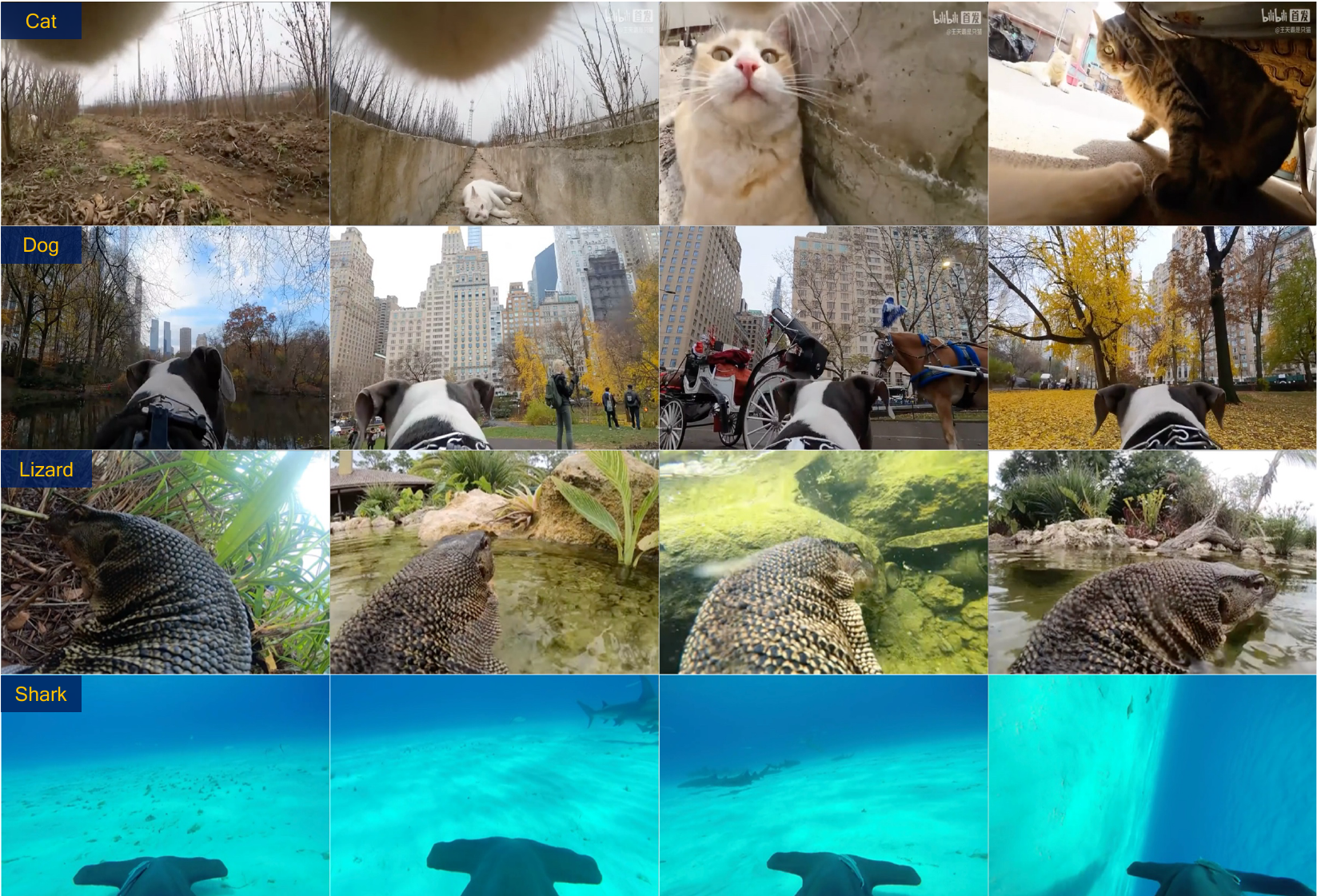}
\caption{\textbf{Additional EgoPet examples}. Footage of four different animal videos from an egocentric view are included.}
\label{fig:supp_dataset}
\end{figure}

We include more dataset visualizations in Figure~\ref{fig:supp_dataset}.


\label{sec:suppexps}

\subsection*{VIP Annotations}
\begin{wraptable}{r}{0.43\textwidth}
\centering
\resizebox{0.85\linewidth}{!}{%
\begin{tabular}{llcccc}
\hline
Model & Dataset & \multicolumn{2}{c}{$t + 4$}\\
& & ATE & RPE \\ 
\hline
MAE & IN-1k & 0.617 & 0.233 \\
MVP & Ego Mix & 0.598 & 0.233 \\
DINO & IN-1k & 0.582 & 0.229 \\
iBOT & IN-1k & 0.574 & 0.226 \\
\hline
VideoMAE & K400 &  0.478 & 0.171 \\
MVD & K400 & 0.479 & 0.172 \\
MVD & Ego4D & 0.477 & 0.172 \\
MVD & EgoPet & \textbf{0.474} & \textbf{0.171} \\
\hline
\end{tabular}
}
\caption{\textbf{Locomotion prediction linear probing results.} Models are evaluated on their ability to predict the trajectory of the agent $k$ seconds into the future.}
\label{tab:planning}
\vspace{1mm}
\end{wraptable}%
We provide additional information regarding the annotation process for the VIP task. The data labeling process for marking interactions involved a meticulous analysis of the video content, which resulted in the annotation of $1449$ subsegments (see Figure~\ref{fig:vip_fig}). Two human annotators were trained to identify and timestamp the start and end of an interaction event. The outcome of this process is a richly annotated dataset of $805$ subsegments where no interaction occurs (''negative subsegments'') and $644$ positive interaction subsegments that capture a wide range of $17$ distinct interaction objects such as person, cat, and dog. The subsegments were then split into train and test. This leaves us with $754$ training subsegments and $695$ test subsegments for a total of $1{\small,}449$ annotated subsegments. 

The beginning of an interaction is marked at the first time-step where the agent begins to give attention to a target, and the endpoint is marked at the last time-step before the attention ceases. In addition, annotators were instructed to mark some segments without interactions. This process results in a set of temporal segments, each corresponding to a discrete interaction event or no interaction event. To ensure the consistency of annotations across annotators, annotations are only kept where annotators agree.

The outcome of this process is a richly annotated dataset of $805$ subsegments where no interaction occurs (''negative subsegments'') and $644$ positive interaction subsegments that capture a wide range of $17$ distinct interaction objects such as person, cat, and dog. The subsegments were then split into train and test. This leaves us with $754$ training subsegments and $695$ test subsegments for a total of $1{\small,}449$ annotated subsegments.

This is the list of all possible interaction objects: Person, Ball, Bench, Bird, Dog, Cat, Other Animal, Toy, Door, Floor, Food, Plant, Filament, Plastic, Water, Vehicle, Other. 

\section*{Results}

\textbf{Full LP results.} Table~\ref{tab:planning} contains the quantitative LP results for all models, reported using the ATE and RPE metrics. The results indicate that pretraining on EgoPet leads to better ATE and RPE scores.

\vspace{1.3mm}
\noindent\textbf{Full VPP results.} In the main paper we included the VPP results grouped by ``past'', ``present'' and ``future'' (see Table~\ref{tab:vpp_res}). In Table~\ref{tab:vpp_supp} we provide the full fine-grained VPP results by individual timestep.

\begin{table*}[t]
\centering
\begin{tabular}{llccccccc}
\hline
Model & Dataset & \multicolumn{2}{c}{Past} & Present & \multicolumn{2}{c}{Future} & Avg. \\
 & & $t-1.5$ & $t-0.8$ & $t$ & $t+0.8$ & $t+1.5$ &\\
\hline
\small{\textbf{1 Frame}} \\
MAE & IN-1k & 0.378 & 0.341 & 0.280 & 0.311 & 0.317 & 0.325 \\
MVP & Egocentric & 0.372 & 0.341 &0.273 & 0.303 & 0.313 & 0.320 \\
DINO & IN-1k & 0.371 & 0.337 & 0.275 & 0.301 & 0.308 & 0.318 \\
iBOT & IN-1k & 0.364 & 0.336 & 0.278 & 0.303 & 0.305 & 0.317\\
\hline

\small{\textbf{2 Frames}} \\
MVD & K400 & 0.333 & 0.284 & 0.207 & 0.255 & 0.273 & 0.270\\
MVD & Ego4D & 0.331 & 0.285 & 0.211 & 0.254 & 0.272 & 0.271
 \\
MVD & EgoPet & 0.328 & 0.281 & 0.197 & 0.248 & 0.271 & 0.265 \\

\hline
\small{\textbf{4 Frames}} \\
MVD & K400 & 0.358 & 0.214 & 0.197 & 0.262 & 0.263 & 0.259\\
MVD & Ego4D & 0.311 & {0.212} & 0.224 & 0.235 & 0.287 & 0.254 \\
MVD & EgoPet & {0.276} & 0.235 & 0.203 & {0.230} & {0.262} & {0.241} \\
\hline
\small{\textbf{8 Frames}} \\
MVD & K400 & 0.226 & 0.208 & 0.196 & 0.240 & 0.264 & 0.227 \\
MVD & Ego4D & 0.217 & 0.200 & 0.192 & 0.234 & 0.264 & 0.221 \\
MVD & EgoPet & 0.214 & 0.195 & 0.184 & 0.237 & 0.268 & 0.219 \\


\hline

\end{tabular}
\caption{\textbf{Vision to Proprioception Prediction (VPP) linear probing results}. We report the mean squared error loss. Models trained on EgoPet perform better than models trained on other datasets. See Supplementary Table~\ref{tab:vpp_supp} for the full  results.}
\label{tab:vpp_supp}
\end{table*}




\section*{Figure credits}
Some of the figures in the paper were created from web videos. We credit the original content creators and provide links to the original videos below.

Figure~\ref{fig:teaser}
\begin{itemize}
    \item \url{https://www.youtube.com/watch?v=69AXB6aFzRU}
    \item \url{https://www.youtube.com/watch?v=0KUhN_rkpAo}
    \item \url{https://www.youtube.com/watch?v=2I6hDqCrI9o}
    \item \url{https://www.youtube.com/watch?v=ZiF5r6d-Ag0}
    \item \url{https://www.youtube.com/watch?v=kpLhZyE37p4}
\end{itemize}
Figure~\ref{fig:dataset-examples}
\begin{itemize}
    \item \url{https://www.youtube.com/watch?v=ALoIn4YpwIc}
    \item \url{https://www.youtube.com/watch?v=ZiF5r6d-Ag0}
    \item \url{https://www.youtube.com/watch?v=qYuAy4I8KDw}
    \item \url{https://www.youtube.com/watch?v=2I6hDqCrI9o}
\end{itemize}
Figure~\ref{fig:VIP_2}
\begin{itemize}
    \item \url{https://www.youtube.com/watch?v=5KTJGuYt2ow}
    \item \url{https://www.youtube.com/watch?v=9dngVyua5_Q}
\end{itemize}
Figure~\ref{fig:lp_2}
\begin{itemize}
    \item \url{https://www.youtube.com/watch?v=g-smZSI0n5Q}
\end{itemize}
Figure~\ref{fig:gradcam}
\begin{itemize}
    \item \url{https://www.youtube.com/watch?v=9A-9Kdmg2C8}
    \item \url{https://www.youtube.com/watch?v=fpcKsAIxtRM}
    \item \url{https://www.tiktok.com/@gonzoisacat/video/7232306745660509483}
    \item \url{https://www.youtube.com/watch?v=EZY7U9faq7g}
\end{itemize}
Figure~\ref{fig:supp_dataset}
\begin{itemize}
    \item \url{https://www.youtube.com/watch?v=WHHmIfog0Fs}
    \item \url{https://www.youtube.com/watch?v=0KUhN_rkpAo}
    \item \url{https://www.youtube.com/watch?v=0t91LFjETis}
    \item \url{https://www.youtube.com/watch?v=auUxD3Qe3FU}
\end{itemize}

\label{suppl:credits}

\label{sec:vip_details}

\end{document}